# Writing Order Recovery in Complex and Long Static Handwriting

Moises Diaz, Gioele Crispo, Antonio Parziale, Angelo Marcelli, Miguel A. Ferrer

## Abstract

The order in which the trajectory is executed is a powerful source of information for recognizers. However, there is still no general approach for recovering the trajectory of complex and long handwriting from static images. Complex specimens can result in multiple pen-downs and in a high number of trajectory crossings yielding agglomerations of pixels (also known as clusters). While the scientific literature describes a wide range of approaches for recovering the writing order in handwriting, these approaches nevertheless lack a common evaluation metric. In this paper, we introduce a new system to estimate the order recovery of thinned static trajectories, which allows to effectively resolve the clusters and select the order of the executed pen- downs. We evaluate how knowing the starting points of the pen-downs affects the quality of the recovered writing. Once the stability and sensitivity of the system is analyzed, we describe a series of experiments with three publicly available databases, showing competitive results in all cases. We expect the proposed system, whose code is made publicly available to the research community, to reduce potential confusion when the order of complex trajectories are recovered, and this will in turn make the trajectories recovered to be viable for further applications, such as velocity estimation.

## 1. Introduction

Over the last 40 years, handwriting analysis and recognition have been widely studied, and many theoretical and experimental results have been obtained both on handwriting acquired with tablets (on-line samples) and on that obtained with scanners (off-line samples) [1], [2]. These studies have contributed to many useful applications, such as mail sorting, form processing and handwriting recognition. More recently, the widespread use of devices such as smartphones, tablets, and electronic pen pads has given rise to a personal digital bodyguard concept [3]. This feature can supplement data protection, which enhances human-machine interactions through handwriting recognition.

In addition to such technological advances, the automatic processing of off-line handwriting is still of great interest in many fields of application, such as enterprise management [4],[5], education [6]– [8], and healthcare [9], [10]. Both public offices and private companies need to archive and retrieve digital versions of documents that contain handwritten samples. Recent years have witnessed the rise of many digital libraries that require systems that allow automatic searches in transcripts and ancient manuscripts [11].

It has long been understood that on-line handwriting systems perform much better than off-line ones because the former have access to dynamic information. This information consists mainly of the order in which the trajectory was written and its velocity profile. This limitation has motivated the development of systems for writing order recovery in static images [12]–[14]. These systems include two steps:

1) thinning of the handwriting, and 2) writing order recovering of the trajectory. Computer-based systems for recovering static trajectories have been proposed for several handwritten applications, including the reading of cursive handwriting [15], Latin and Arabic handwriting word recognition [16]–[18], Indian and Chinese character recognition [19], [20], digit recognition [21], historical document transcription[22], mathematical symbol recognition [23], signature verification [14], [24], [25], writer verification and identification [26], [27], handwriting style modeling [28], [29] and handwriting analysis and synthesis [30].

Recovering the trajectory in static handwriting may lead to multiple possible solutions, with only one representing the real trajectory sought. Solving this inverse problem becomes even more complicated when the handwritten pattern contained in the image comprises multiple components;[1] in this case, recovering a new component may depend on the order of the previous one and the in-air trajectory between two consecutive components.

---

[1] *By "component", we refer to a piece of writing with two end-points, meaning that it is performed between two pen-ups. It can be denoted as a pen-down as well.*

### A. Our Contribution

This paper proposes a novel system for estimating the writing order recovering in complex and long static handwriting consisting of many components, separated by pen-ups, also named in-air trajectories. Inspired by human movements during the production of handwriting and by motor control perspectives [31], our system chooses the smoothest ballistic trajectories based on good continuity criteria when writing or drawing. To this end, we apply some multiscale strategies to recover the trajectory order in the agglomeration of pixels, which we call clusters. Additionally, we study the proposed system's performance, both when the end-points of the components are known and when they are unknown. We provide a quantitative measure of how heuristics on the choice of starting and ending points affect the writing order recovery. The effectiveness of the system is demonstrated in three databases, achieving competitive performance.

We focus our work on static specimens for two reasons: the first one is that developing a system that recovers complex, long, and discontinuous trajectories represent the most challenging case, and the second is that estimating the writing order from static images is of great interest for the early prescreening of neurodegenerative disorders [10], [32].

The code is developed in Matlab, and it is freely provided for research[2].

The outline of the paper is as follows: Section II presents a brief overview of related works on writing order recovery of trajectories in static handwritten specimens. Section III describes the proposed solution for estimating handwriting recovering. A sensitivity and stability study of parameters is given in Section IV, whereas experimental results are presented in Section V. Finally, conclusions are drawn in Section VI.

## 2. Literature Review

Over the last thirty years, many systems have been proposed for recovering static trajectories, with two surveys assessing the related state of the art [12], [13]. Moreover, a competition has been considered in order to establish a common benchmark for writing order recovery of Arabic signatures [14], [33].

Systems used to recover trajectories from static handwriting employ three main approaches: contour-based, skeleton-based, and learning-based. The first two differ in terms of the "object" used to represent the handwriting, respectively its contour and its skeleton. The skeleton is the result of a thinning process that produces a 1-pixel- wide line, which follows the centerline of the original image, and that ideally corresponds to the original pen-tip trajectory.

Regarding contour-based approaches, [34] describes a method that segments cursive handwriting by detecting the points where the trace contour has the maximum curvature. The segments are ordered based on the contour curvature smoothness. Another handwriting segmentation is given in [16] and proceeds through an analysis of the handwritten contour. Here, the writing order trajectory is estimated by adopting graph-based representations at both the segment and stroke levels. The list of candidate paths is obtained by choosing curvature and width stroke preservation, which are local continuity criteria, as cost functions. Following up on [16], the system was employed in [17] to develop an off-line recognition system. In that work, a Hidden Markov Model (HMM) was trained for selecting the most likely written trajectory from the list of candidate paths. To avoid introducing artifacts in the handwriting, preprocessing steps, such as binarization, were omitted in [35]. Instead, the authors extracted control points from the grayscale images, which were used to recover the trajectory according to certain heuristics rules.

The skeleton-based methods can be categorized as local line order recovering and global graph searching methods. The former reconstructs the trajectory by choosing the most plausible direction at each branch point of the skeleton. These methods are simple and have a low computational cost, but their performances are typically limited by the difficulty of designing heuristic rules for different handwriting styles. Instead, graph searching methods recover the trajectory by representing the topological structure of the skeleton with a graph and traversing it. They have a greater computational cost and their performances depend on the definition of criteria for selecting the best trajectory among many alternatives.

An example of local line recovering is given in [36], where the trajectories are recovered according to good continuity criteria, which take into account the direction, length, and width of the strokes making up characters. Off-line signatures are recovered in [37] by following heuristic rules inspired by the way human beings write a given shape. The rules are applied to deal with low-level pixel processing and high- level stroke processing. A similar approach is applied to handwritten digits in [38]. In [39], a taxonomy of local, regional, and global features that can be used for recovering temporal properties from the image is proposed.

---

On the other hand, a likelihood measure is developed in [40]. It selects the most likely writing order recovery from the analysis of skeletons. In an on-line automatic signature verification system, the method proposed in [40] shows a greater false acceptance rate in skilled forgeries than in random forgeries. A genetic algorithm is also used for recovering segments extracted from the skeleton [41]. The fitness function adopted by the genetic algorithm for selecting the best individual of the population takes the writing direction, the repetition of segments, and the angular deviation on the crossing of the occlusion stroke into account. Additionally, a complete framework to recover the dynamic properties (i.e., velocity and pressure) from an image-based signature is presented in [24] by using classical approaches to recover the static trajectories in the signatures.

As for global graph methods, they represent the topological structure of the skeleton with a graph whose vertices represent the end-points, the junctions and the contact points of the skeleton, and whose arcs represent lines and curves. These methods determine the writing order recovery by finding the most appropriate path along the graph. In [42], one of these methods is proposed for recovering the trajectory of words through the traveling salesman problem. Based on the skeleton of words, the authors search for trajectories with minimum curvatures. Also, in [43], each component of a word is represented by a graph. The trajectory of the whole word is obtained by concatenating, from left to right, the most likely trajectory of each component. The authors define both global and local criteria that allow to select the best trajectory. The path search was based on the best-first search algorithm The authors in [44] propose to construct a graph to use to determine the types of each edge from the skeletons. Then, they develop a writing order recovery algorithm to traverse such a graph without applying any graph search algorithm. The goal of the method proposed in [45] is to use as little heuristic knowledge as possible. To this end, the proposal applies the maximum weighted matching of general graphs to find double-written lines. It exploits the minimum energy cost criterion as a guiding principle for recovering the trajectories.

Successful results in single-stroke images are obtained.

A graph transformation is proposed in [46] to ensure that all graphs' nodes had an even number of incident arcs. This property allows to traverse the graph by using Fleury's algorithm, combined with handwriting generation models. The approach exhibited a reduced computational cost because it divided the whole graph into sub-graphs by detecting the graphs' bridges. This method is improved in [47] by introducing a feedback connection between the unfolding module and a module that extracts elementary movements from the recovered trajectory. The method exploits the analogies between unfolding and segmentation processes and those occurring in the brain when a trajectory plan is learned and executed.

Furthermore, the search for the optimal trajectory along the graph representing the skeleton of a word is executed in [48]. The authors use both the Greedy and the Dijkstra algorithms with a well-defined smoothness function. Further, they focus on reconstructing the optimal trajectory of words by splitting them into many strokes according to their respective curvature values and a set of rules of handwriting. Good continuity criteria derived from both visual perception and movement execution are applied to signatures in [50]. In particular, the implementation focuses on a multiscale analysis of the thinned trajectories and the Dijkstra algorithm.

For learning-based approaches, they require some exemplars of static images with the corresponding drawing orders. Thus, models are trained to recover the trajectories of new static images. In [51], an HMM is adopted for recovering the trajectory of single-stroke handwritten signatures. Each state of the HMM has a probability density function that embeds geometric shape information of the static image, while transition probabilities define the possible pen movements between static image coordinates. A training phase is proposed in [52], in which the original trajectory order and other attributes, such as the length and direction, are extracted from a set of on-line scripts to build a universal writing model. Then, it is used to reconstruct the drawing order during the test phase. The skeleton of the static image is matched to the model by using a dynamic programming algorithm, and the trajectory with maximum likelihood is selected.

Since 2018, deep learning techniques have been exploited in recovering trajectories in Chinese, Japanese, Indic, Arabic, and Latin characters (e.g. [41], [58]). While these techniques may be promising, they have the disadvantage of requiring a huge amount of data for training. Presently, deep learning methods are used to estimate the trajectory of characters and numbers, which are less complex than words and signatures. In [53], an algorithm is proposed based on a regression Convolutional Neural Network (CNN) model to predict the probability of the next stroke point position. The same authors present improvements of such an architecture in [54], based on two CNNs. Another model based on an encoder-decoder LSTM module was introduced in [20].

Here, the encoder module consists of a Convolutional LSTM network, which takes an off-line character image as input and encodes the feature sequence to a hidden representation. The output of the encoder is fed to a decoder LSTM that sequentially predicts the coordinate points. The architecture is tested on characters from three Indic scripts. Experimentation shows that the main limitation of the approach is the need to train a separate model for each individual script.

A handwriting recognition system based on a writing order recovery algorithm is proposed in [55]. The order recovery algorithm exploits an end-to-end system based on a VGG-LSTM, which extracts and encodes features, followed by a BLSTM used as a decoder to generate temporal coordinates. The method could eventually produce human-like velocities [18]. Moreover, a network of two variational auto-encoders is proposed in [56] to convert on-line and off-line handwritten Latin characters to each other. An improved VGG-16 CNN model is proposed in [57] to recover the handwriting stroke order. The CNN model recovers the writing order effectively, even if the accuracy of the network decreases as the number of strokes increases.

As summarized in Table I, an analysis of the state of the art shows that the static handwriting trajectory recovering problem is far from solved despite the very high number of papers that have been published on the subject. Many methods have been designed, and they mostly start from the assumption that the static image consists of a single component, i.e., a pen-down. This assumption does not hold in real applications, where handwriting patterns consist of many pen- downs and pen-ups, such as in signatures. Furthermore, many systems on the subject have not been validated on public datasets, but rather, have only been tested on a few samples. Even more, performance of some systems are only qualitatively measured. Eventually, we notice that studies reported in Table I adopt different metrics and strategies for measuring the goodness of the writing order recovery systems: some authors use the accuracy obtained by recognition systems on the recovered trajectory as an indirect metric, whereas others compare the recovered trajectory with the data acquired by a tablet.

## 3. Estimating the Writing Order in Handwriting

The objective of the proposed system is to estimate the writing order of the 8-connected thin line representation of the handwriting image. The analysis of two elements (components and clusters) plays a vital role in the trajectory recovering process. A cluster is formed following the intertwining of different strokes, resulting in an agglomeration of pixels. Each pixel in the cluster therefore has more than two pixels connected in its 8-neighborhood, and consequently, the clusters complicate the accurate component drawing process.

A flowchart of the proposed system is depicted in Fig. 1. It is composed of three stages. 1) *Point classification*, where the clusters of a thinned specimen are identified, 2) *Local examination*, where individual clusters are resolved by joining their output branches, and 3) Global reconstruction, where we estimate the handwriting order with multiple pen-downs. The mathematical notation used in this work is provided in Table II.

### A. Point Classification

We classify each point (or pixel) of the thinned trajectory into one of three categories: (1) end-points, which are pixels with only one 8-neighbor, (2) trace points which have two 8-neighbors, and (3) branch points which have three or more 8-neighbors. Fig. 2 gives a visual example of this categorization. To this end, we identify the clusters as the sets of adjacent pixels labeled as branch points.

To classify the black pixels in the image, we search for their connectivity in their 8-neighbors. The complexity order of this procedure is $\mathcal{O}$ ($8 \cdot h \cdot w$), where ($h, w$) are the height and width of the image, respectively. Next, the clusters are defined as the sets of adjacent pixels previously labeled as branch points. The clusters are thus identified by performing the connected component labeling algorithm over the branch points. As this algorithm requires at most two scans of the image, its computational complexity is $\mathcal{O}$ ($2 \cdot h \cdot w$).

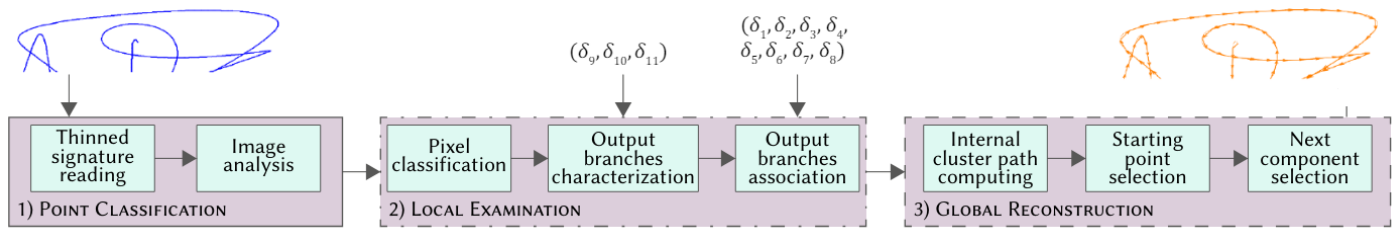

Fig. 1. Flowchart of the proposed system for writing order recovering.

*Table 1 Literature Review on Systems for Recovering Static Trajectories*

| Ref. | Approach | Pattern | Dataset | Evaluation |
|---|---|---|---|---|
| [34] | Contour-based | Words | Private. 200 images of words written by six different writers. | Quality of ordering ranked by human subjects. |
| [16], [17] | Contour-based | Words | Public. 10,448 words taken from the IRONOFF dataset. For each word, the on-line and off-line versions are available. | Accuracy of HRS. |
| [35] | Contour-based | Signatures | Public. SVC 2004 dataset. Images are synthesized from on-line samples. | Preliminary study. The number of detected stroke points. |
| [36] | Skeleton - Local line | Characters | Private. 10,000 characters written 20 subjects. 2/3 of samples used as a test set. | Quality of ordering ranked by human subjects. |
| [37], [38] | Skeleton - Local line | Signatures and Numerals | Private. 20 signatures [37] and 150 numerals [38]. | Visual inspection for signatures and HRS for numerals. |
| [39] | Skeleton - Local line | Words and Numerals | Private. 1000 images in US mail service. | Visual inspection. |
| [40] | Skeleton - Local line | Signatures | Public. The last fifty users of the MCYT-100 dataset wrote the signatures. Images are synthe-sized from on-line samples. | FAR and FRR of ASV. |
| [41] | Skeleton - Local line | Words | Public. Hundreds of samples from LMCA, IRONOFF, and IFN/ENIT datasets. | Visual inspection. |
| [24] | Skeleton - Local line | Signatures | Public. Fifty users of the BiosecurID dataset. Images are synthesized from on-line samples. | EER of ASV. |
| [42] | Skeleton - Graph-based | Characters | Private. | Visual inspection. |
| [43] | Skeleton - Graph-based | Words | Private. 150 words written by five subjects. | Accuracy of HRS. |
| [44] | Skeleton - Graph-based | Words | Private. 100 images. | Visual inspection. |
| [45] | Skeleton - Graph-based | Characters | Public. 708, 811 images obtained by converting on-line data of the Unipen dataset. | Rate of correct complete trajectory recovery, Ac-curacy of HRS. |
| [46],[47] | Skeleton - Graph-based | Words | Private. 6500 images containing cursive handwrit-ing. | Similarity measure between automatic and man-ual drawing order. |
| [48] | Skeleton - Graph-based | Words and Characters | Private. 6868 images taken from 3 different datasets. Pubic. Images were taken from IRONOFF [49]. | RMSE, DTW, Accuracy of HRS. |
| [50] | Skeleton - Graph-based | Signatures | Public. 1953 and 2820 on-line signatures from the SigComp2009 and SUSIG - Visual datasets, respectively. The thinned version of the on-line signatures is used. | RMSE, DTW, Number of clusters correctly solved. |
| [51] | Learning-based | Signatures | Public. 710 single-stroke on-line signatures from fifty users taken from US_SIGBASE and Dolfing datasets. Images are synthesized from on-line samples. | Accuracy score that measures the alignment of the recovered order and the ground truth. |
| [52] | Learning-based | Signatures | Private. 300 images of signatures. | The rank of the proposed recovering trajectories. |
| [53], [54] | Learning-based | Characters and digits | Public. OLHWDB 1.1 dataset containing 3755 Chinese characters; 2000 English letters and Arabic digit symbols in UNIPEN. | Rate of correct: end point selection [54], branch points resolution [54], complete trajectory recov-ery [53], [54]. |
| [20] | Learning-based | Characters | Public. LIPI Toolkit dataset (Tamil, Telugu and De-vanagari characters). Around 21, 000 characters per script. | Rate of correct: starting point selection, junction points resolution, and complete trajectory recovery. |
| [55] | Learning-based | Characters and digits | Public. LMCA and IRONOFF datasets. | Accuracy of HRS. |
| [56] | Learning-based | Characters | Public. Unipen. | DTW. |
| [57] | Learning-based | Characters | Public. OLHWDB 1.1 dataset. | Rate of correct complete trajectory recovery. |

HRS: Handwriting Recognition System, ASV: Automatic Signature Verifier.

*Table 2. Notation used*

| Notation | Description |
|---|---|
| $r$ | Rank of the cluster |
| $c_{ij}$ | Curvature between output branches i and j |
| $\alpha_i, \beta_i$ | External and internal angle for branch i |
| $\pi_{ij}$ | Weighted angle direction between branches i and j |
| $\omega_{ext}, \omega_{int}, \omega_{cur}$ | External, internal and curvature weights for computing the anchor point direction |
| $\delta_1, \delta_2, \delta_3$ | Parameters for identifying retracing |
| $\delta_4, \delta_5, \delta_6$ | Parameters for identifying T-patterns |
| $\delta_7, \delta_8$ | Parameters for identifying coupled clusters |
| $\delta_9$ | Number of trace points |
| $\delta_{10}$ | Parameter for the brotherhood |
| $\delta_{11}$ | Number of points for computing the curvature |

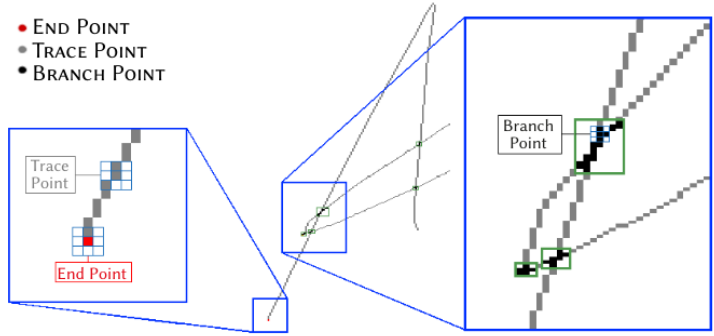

*Fig. 2. Point classification in a thinned trajectory (center of the image): details inside the blue rectangles with clusters inside the green ones.*

### B. Local Examination on the Clusters

Recovering a skeleton from an end-point to another through the trace points is an effortless operation. However, the task becomes more demanding when clusters are encountered, since we must then decide on the adequate output branch to recover. The output branches can be also defined as all the trace points that converge in a single cluster. For this writing order recovery, given a cluster, firstly, its pixels are classified. Secondly, the output branches are characterized, and, finally, they are paired off in input-to-output paths.

### C. Pixel Classification in a Cluster

The pixels inside a cluster are classified according to their connectivity as follows:
• Cluster points. These are branch points.
• Anchor points. These are branch points of the clusters having at least one trace point (i.e., a point outside the cluster) as a neighbor. Since the cluster output branches are anchored on them, they are denoted as anchor points. Moreover, the number of anchor points establishes the rank of the cluster, denoted by r.
• False trace points. These are labeled as cluster points whether at least one of the following conditions hold: (1) they are connected to two different cluster points of the same cluster; (2) they are connected to a cluster point and to a false trace point of the same cluster; (3) they are connected to two false trace points of the same cluster. Such a classification in a single cluster requires a computational complexity of $\mathcal{O}(p2)$, with p being the total number of cluster points. Fig. 3 shows an example of a cluster and its classified pixels.

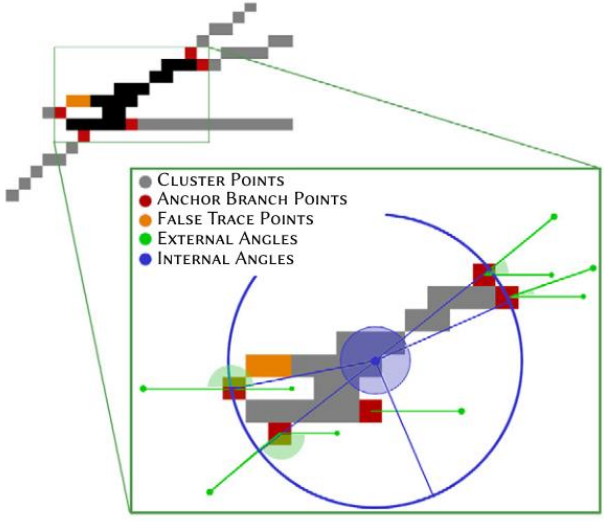

*Fig. 3. Example of a cluster: cluster points (in gray), anchor points (in red), false trace points (in orange), external angle directions (in green), and internal angle directions (in blue).*

## D. Output Branches Characterization

Cluster points and false trace points are identified through their horizontal and vertical coordinates in the images. Beyond the Cartesian coordinates, the anchor points are characterized through the angle direction of the output branches. We developed three strategies for computing the direction of the output branches.

**- External angle**. A multiscale approach [59] is devised to ompute this angle by using only information external to the cluster. This algorithm considers the $\{x_i, y_i\}_{i=1}^{i=\delta_9}$ coordinates of an anchor point $(AP_x, AP_y)$ and $\delta_9$ numbers of trace points of the output branch (see Table II). The output of this algorithm is the external angle $\alpha_i$, $\forall i \in 1, ..., r$. The algorithm is formalized in Algorithm 1, which runs in time $\mathcal{O}(\delta_9^2)$. Additionally, Fig. 3 illustrates the external angles on the anchor points.

**- Internal angle**. The internal angles can compensate for some shortcomings of the external angles when the output branches contain a few trace points. For this compensation, the cluster center of gravity of the anchor points is calculated. The $(x, y)$ coordinates of the first $\delta_9$ trace points from an output branch are considered and the internal angles $\beta_i$, $\forall i \in 1, ..., r$ are computed by the procedure reported in Algorithm 2, its complexity being $\mathcal{O}(\delta_9^2)$.

**- Curvature**. We combine all pairs of output branches, $C_2^r = \binom{r}{2}$, chosen among the $r$ available anchor points in a cluster. The continuity connection of two output branches $(i, j)$ and its shortest path is obtained through Dijkstra's algorithm [60]. Each of the 8-connected traces thus created is divided into $n$ equidistant points and a curvature representative value is calculated $(\rho = \rho_1, \rho_2, ..., \rho_n])$. Finally, the maximum of the difference between adjacent elements of $\rho$ is used to quantify the curvature of the trace generated with the output branches $(i, j)$. The curvature $c_{i,j}$ goes from 0, if the curve is perfectly straight, to 180 degrees, if the curve is bent over itself. The procedure is detailed in Algorithm 3, which has a complexity of $\mathcal{O}(\delta_{11}^2)$.

## Algorithm 1. Compute External Angle
1: procedure computeExtAng($AP_x$, $AP_y$, x, y)
▷ $\delta_9$ being the scale of the multiscale approach and the length of (x, y)

2: for s = 1 to $\delta_9$ do
3: k ← 1
4: nextPixel ← ($AP_x$ , $AP_y$ )
5: for i = 1 step s to $\delta_9$ do
6: $dist_x$ ← $nextPixel_1 - x_i$
7: $dist_x$ ← $nextPixel_2 - y_i$
8: $angInt_k$ ← atan2($dist_Y$, $dist_X$)
9: k ← k +1
10: nextPixel ← ($x_i$ , $y_i$ )
11: end for
12: minAngle ← min of angInt
13: maxAngle ← max of angInt
14: $angMS_{(s,:)}$ ← linear interpolation
15: end for
16: for s = 1 to $\delta_9$ d
17: resMultiscale ← circular mean of $angMS_{(s,:)}$
18: end for
19: α ← circular mean of resMultiscale
20: return α

### E. Handling Special Cases: the Brotherhood

The brotherhood refers to a set of joined clusters. Two clusters can be merged when there are less than $\delta$ trace points from each other and are connected by at least two branches. The brotherhood is a full-fledged cluster, and we apply the same rules of a single cluster to it. Fig. 4 shows two examples of brotherhood process. It is worth pointing out that cluster C3 (Fig. 4a) and cluster C1 (Fig. 4b) are not included in the brotherhood because they are connected with the other clusters only through one branch.

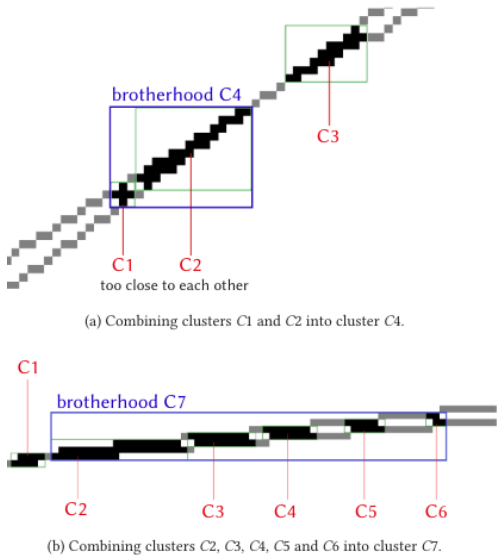

(a) Combining clusters *C1* and *C2* into cluster *C4*.

(b) Combining clusters *C2*, *C3*, *C4*, *C5* and *C6* into cluster *C7*.

*Fig. 4. Examples of two brotherhood application.*

The complexity of the brotherhood is $\mathcal{O}\left(n\mathsf{C} \cdot r \cdot \delta_{10}^2\right)$., where $n\mathsf{C}$ is the number of clusters in the brotherhood, r the rank of the cluster and $\delta_{10}$ the number of trace points.

**Algorithm 2.** Compute Internal Angle
1:  procedure computeIntAng($ccg_x$ , $ccg$ , $x$, $y$)
▷ $\delta_9$ being the scale of the multiscale approach and the length of ($x$, $y$)
2:      for $s = 1$ to $\delta_9$  do
3: $vx_i$ ← $ccg_x - x_i$

4: $vy_i \leftarrow ccg_y - y_i$
5: $ang_i \leftarrow$ atan2d(vxi , vyi )
6: end for
7: $\beta \leftarrow$ circular mean of ang
8: return $\beta$

**Algorithm 3**. Compute Curvature
1: procedure computeCurv(x, y, $\delta_{11}$)
2: $m \leftarrow$ length of (x, y)
3: $n \leftarrow \min(\delta_{11}, m)$
4: if $m \leq 2$ then
5: return 0 ▷ The curve is straight
6: else
7: $s \leftarrow$ floor(linspace(1, m, n))
▷ Indexes of (x, y) to be considered
8: for $1 = 1$ to n do
9: for $a = 1$ to $\delta_{11}$ do ▷ Foward points
10: if $1 \geq 1$ and $1 \leq n - 1$ then
11: $k \leftarrow \min(n - l, a)$
12: for $f = 1$ to k do
13: $v_f \leftarrow$ atand2 $(y_{s_1} - y_{s_{1+f}} , x_{s_1} - x_{s_{1+f}} )$
14: end for
▷ Backward points
15: if $1 \geq 2$ and $1 \leq n$ then
16: $k \leftarrow \min(l - 1, a)$
17: for $b = 1$ to k do
18: $v_b \leftarrow$ atand2 $(y_{s_1} - y_{s_{1+b}} , x_{s_1} - x_{s_{1+b}} )$
19: end for
20: $v \leftarrow$ concat($v_b$ , $v_f$)
21: $ang_a \leftarrow$ circular mean (v)
22: end for
23: $\rho_l \leftarrow$ circular mean (v)
24: end for
25: $c = \max(\Delta_{\rho_l} )$
26: end c

### F. Output Branches Association

The Gestalt theory [61] states that all elements of sensory input are perceived as belonging to a coherent and continuous whole. Moreover, such criteria are supported by the studies of motor control theories [62], in particular those related to the execution of rapid and smooth movements that involve the principle of energy minimization. Under these perspectives, good continuity criteria are taken into account.

As such, we pair the two exit directions whose external difference is closer to 180 degrees than are the others, remove them from the cluster, and repeat the pairing until the rank of the cluster is either zero (meaning there are no more exit directions to pair) or three. Let i and j be two branches of a cluster. To implement this criterion, the following weighted angle direction, $\pi_{i,j} \forall (i, j) \in r$, is calculated:

$$\pi_{i,j} = \omega_{ext} \cdot |\alpha_i - \alpha_j| + \omega_{int} \cdot |\beta_i - \beta_j| + \omega_{cur} \cdot C_{i,j} \qquad (1)$$

where $(\alpha_i, \alpha_j)$ refer to the external angles, $(\beta_i, \beta_j)$ to the internal angles and $c_{i,j}$ denotes the curvature between the considered branches. The weights in the angles and curvature, $(\omega_{ext}, \omega_{int}, \omega_{cur})$ (see Table II), have to satisfy that the sum of their values be equal to one, with each weight ranging in the $(0, 1)$ interval. According to eq. (1), the smaller $\pi_{i,j}$ the smoother the line connecting the i-th and j-th output branches. Once all the $\pi_{i,j}$ are calculated, we process the clusters depending on their ranks:

• for even-rank clusters, we select the pair corresponding to the smallest values of $\pi_{i,j}$ and removing the paired branches. These steps are repeated $\frac{r}{2}$ times, until all the branches are paired.

• for odd-rank clusters, the same procedure as above is applied $\frac{r-1}{2}$ times. The remaining branches constitute a 3-rank 2 cluster.

The 3-rank clusters are by far the toughest to manage [45],[63]. They can be classified according to their geometrical and morphological properties (see Fig. 5) as follows:

• **T-pattern clusters:** These are clusters whose shape is similar to a "T". With these, one out of the three angles $\pi_{i,j}$ had to satisfy $180\,(1-\delta_4) \le \pi_i, j \le 180\,(1+\delta_4)$, and the other two, the following condition: $\pi_{i,j}\frac{100}{360} \le \delta_5$. To avoid 360 misclassification, any branch of a T-pattern cluster should not be too close to an end-point and to another 3-rank cluster. Let dep be the distance in pixels between an anchor branch point and the nearest end-point and let d3rc be the distance between an anchor branch point and the nearest 3-rank cluster. Then, the last condition to satisfy to consider a T-pattern would imply: min $(\mathrm{d^{ep}}, \mathrm{d^{3rc}}) \ge \delta_6$ .

• **Retraced clusters:** These appear in closed handwritten loops, and consequently, at least one of the branches has to end in an end point. Let $d_k^{ep}$ be the minimum distance in pixels from the branch k to its end-point. Then a retracing should satisfy $d_k^{ep} \le \delta_2$. Also, as the candidate retraced branch is expected to be straight, the value of its curvature ck should be less than or equal to $\delta_3$. Finally, the opposite angle $\pi_{i,j}$ should also satisfy: $\pi_{i,j}\frac{100}{360} \le \delta_1$ .

• **Coupled clusters:** These are two-neighbor 3-rank clusters, none of which is a T-pattern or a retraced cluster, which share one output branch whose length $\mathrm{d_{db}} \le \delta_7$ . As such, the other branches of both clusters compose a 4-rank cluster, which has to respect the following relation:

$$\max\,[\mathrm{avg}\,(\pi_{1,3}, \pi_{2,4}),\, \mathrm{avg}\,(\pi_{2,3}, \pi_{4,1}),\, \mathrm{avg}\,(\pi_{2,1}, \pi_{3,4})] \le \delta_8 \qquad (2)$$

This condition means that the good continuity criterion must be strongly observed for these clusters to be classified as coupled.

• **Normal cluster:** This is a 3-rank cluster that does not belong to the previous classes. In this case, the branches corresponding to the smallest values of $\pi_{i,j}$ are associated, and the remaining one is disjoined from the cluster.

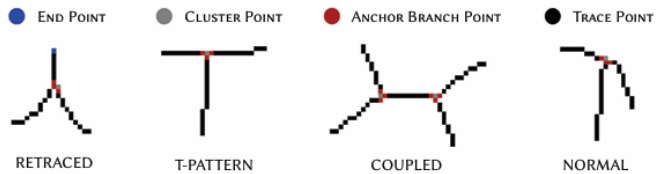

*Fig. 5. Classification of the 3-rank clusters.*

In our implementation, firstly, we pair the branches in even rank clusters, because these clusters are the less ambiguous ones. Secondly, we work out the odd cluster whose rank is higher than three. Finally, we deal with the three rank clusters since they are the most confusing type.

### G. Global Reconstruction

The goal of this section is to recover the writing order of the pen downs. To this end, we need to define a path that traverses the cluster to link two previously paired branches. Moreover, the starting points of each component and their order need to be estimated as well.

*H. Computing Internal Cluster Paths*

Once the anchor branch points are paired, we define the path that connects them. Let a cluster be composed of p pixels. The adjacency matrix A is then a p × p matrix defined as follows: Given a pair of neighbor pixels $(p_i, p_j)$ within the cluster, A(i, j) would have a value of two or three. If $(p_i, p_j)$ are not neighbors or $p_i = p_j$, the value in the adjacency matrix is zero. Based on observations in trajectory generation [31], the weights assigned operate in compliance with the principle of minimizing energy. Accordingly, to connect two neighboring pixels, we exercise a preference in choosing a straight connection rather than an oblique one. At the same time, an oblique connection is preferred against two straight connections.

The adjacency matrix A is therefore processed by the Dijkstra algorithm [60], which approaches the Gestalt theory perspective of good continuity [61] and rapid movement trajectories.

*I. Starting Point Selection*

To choose the starting point, we model the spatial distribution of the starting points with a two-dimensional Gaussian function.

To this end, we store the starting end-point of a random number of handwriting. The median of this Gaussian function is experimentally located at 0.15 · h and 0.35 · w, with h and w being the height and the width of the writing area on the top-left part of the images. When there are no end-points within the ellipse defined by the mean and the two standard deviations of the Gaussian, the leftmost end-point is selected as the starting point, as usual in Western handwriting.

*J. Next Component Selection*

Once the first component is recovered, the next component is chosen according to a proximity criterion; this is the component with the nearest end-point, which has not yet been recovered. Formally, this criterion can be described as:

$$i^* = \arg\min_{i \setminus j} \left( \sqrt{\left(x_{ep_i} - x_{ep_j}\right)^2 + \left(y_{ep_i} - y_{ep_j}\right)^2} \right)$$

(3)

where $(x_{epi}, y_{epi})$ are the coordinates of the last recovered end-point and $(x_{epj}, y_{epj})$ the coordinates of the end-points of components that have not yet been recovered.

Finally, Fig. 6 illustrates three examples obtained with the proposed method.

*IV. Experimental Sensitivity and Stability Analysis*

In this section, we analyze the sensitivity and the stability of our system in terms of the values of the parameters listed in Table II, $\delta_1$ to $\delta_{11}$, $\omega_{ext}$, $\omega_{int}$, and $\omega_{cur}$, which have been determined heuristically. The performance of the system depends on the parameter values of the system. We define the accuracy rate θ as a performance measure:

$$\theta = \frac{\#clusters\ correctly\ solved}{\#total\ clusters}$$

A cluster is counted as correctly solved if all its branches are paired similarly to the real trajectories, which are used as ground truth data.

The parameters and weights were heuristically optimized in a trial and error procedure in three steps. Initially, all the weights were set to 0.33 and a coarse tuning of the parameters $\delta_k$ was conducted.

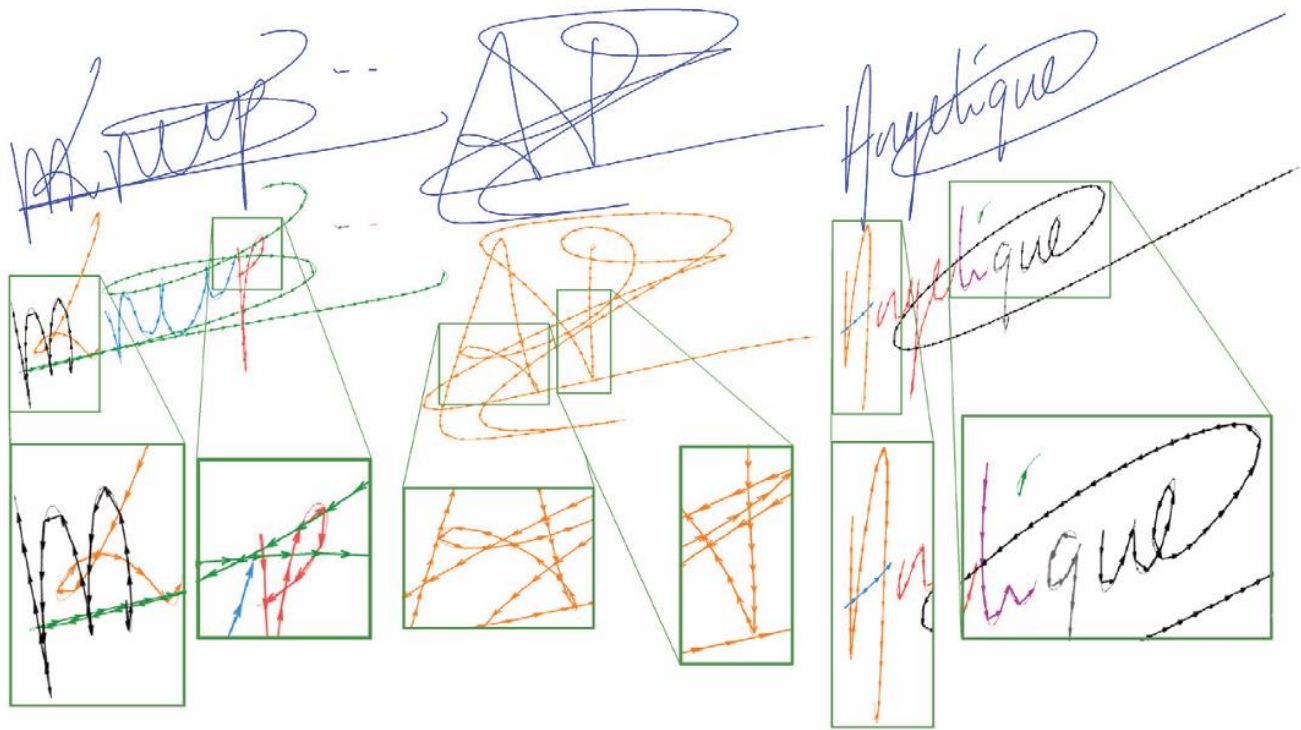

*Fig. 6. Examples of writing order recovering in signatures with the proposed method. The recovered writing order of each component is represented by directional arrows, whereas different colors in the estimated trajectories represent different detected components.*

Then, the performance of θ was analyzed independently for each decision criterion, cluster rank and type to fine-tune the parameters. Eventually, the weights were fine-tuned for a final maximization of θ.

To investigate whether the heuristic optimization of parameters was biased, we carried out both a sensitivity and a stability study. In both cases, our optimization procedure was performed with 30 % of the specimens randomly selected from the SigComp2009 database [64]. Each experiment was repeated ten times with different values of each parameter under investigation, and the corresponding average accuracy rates were reported.

### A. Sensitivity Study:

Results In the study, the heuristic values of the parameters $\delta_k$ were individually varied, and the results given in terms of the sensitivity grade, defined as $\Delta\theta/\Delta\delta$.

*TABLE III. Values of $\delta_k$ Parameters and their Variability Range for the Sensitivity Study*

| $\delta_k$ | Value | Type | Range |
|---|---|---|---|
| $\delta_1$ | 28 | ND* | (22, 34) |
| $\delta_2$ | 20 | ND | (10, 30) |
| $\delta_3$ | 20 | degrees | (10, 30) |
| $\delta_4$ | 3 | ND | (1, 5) |
| $\delta_5$ | 19 | ND | (16, 22) |
| $\delta_6$ | 8 | Pixels | (4, 12) |
| $\delta_7$ | 50 | Pixels | (46, 56) |
| $\delta_8$ | 40 | ND | (20, 60) |
| $\delta_9$ | 5 | Pixels | (3, 7) |
| $\delta_{10}$ | 10 | Pixels | (6, 14) |
| $\delta_{11}$ | 10 | Pixels | (6, 14) |

*ND stand for non-dimensional.

Table III shows the heuristic values and the variation range we used for each of them. For example, the parameter $\delta_9$ had to include at least two pixels, and as a result, its lowest limit had to be 3.

The sensitivity grade for each parameter variation is shown in Fig. 7. It refers to the accuracy rate variation achieved when only one parameter varies, while the remaining ones assume their heuristic values. We see that, depending on the parameter, the variation of the accuracy rate is in the order of magnitude $10-3$ and can therefore be considered negligible. The results confirm that the heuristic criteria we designed, based on the Gestalt theory of perception and motor control theory, capture some properties of the writer's movements when their trajectories exhibit intersections or crossings. We can also see that $\delta_3$ exhibits the highest sensitivity grade, meaning that it is one of the most influential parameters, and conversely, $\delta_6$ seems to be the least influential one.

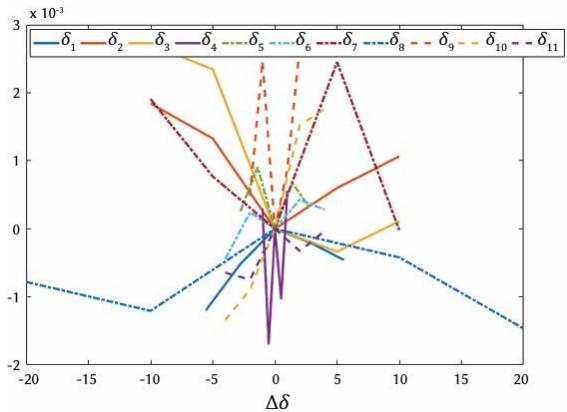

*Fig. 7. Sensitivity analysis: Each line shows the variation of the accuracy rate $\Delta\theta$ for a different variation $\Delta\delta$ of the parameter $\delta_k$.*

### B. Stability Study: Results

In this study, different weight values are obtained by adding Gaussian noise to the heuristic values, as:
$\hat{\omega} = \mathcal{N}(\omega, (\eta \cdot \omega)^2)$ where $\eta \in (0.05, ..., 0.5)$ is a distortion factor of the standard deviation of the Gaussian. In each experiment, the value of $\eta$ was incremented by $0.05$ and normalization was carried out in order to satisfy $|\omega_{ext}| + |\omega_{int}| + |\omega_{cur}| = 1$.

Table IV gives the heuristic values of the weights, whereas Fig. 8 shows the accuracy rate obtained for different distortion levels. For each value of $\eta$ the figure reports the corresponding box plots. As it can be seen, the central mark, in red, is above $\theta = 0.95$ and the majority of data, in blue, are concentrated around the median. Black dots represent the outliers in each case. Consequently, highly stable performances are obtained, along with different $\eta$ values. It suggested that external and internal angles, as well as the curvature conditions, are representative of the good continuity criteria.

*TABLE IV. Weights for Computing the Branch Point Directions*

| Clusters | $\omega_{ext}$ | $\omega_{int}$ | $\omega_{cur}$ |
|---|---|---|---|
| Normal | 0.20 | 0.05 | 0.75 |
| T-Pattern/Retracting | 0.95 | 0.00 | 0.05 |
| Coupled | 0.40 | 0.05 | 0.55 |
| Odd-Rank | 0.70 | 0.05 | 0.25 |

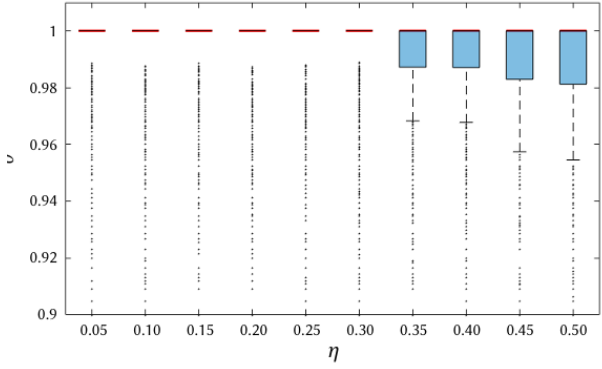

*Fig. 8. Stability analysis: Vertical axis shows the accuracy rate (θ) as a function of the amount of distortion (η) applied to the heuristic values of the weights, in the horizontal axis.*

## 5. Performance Assessment

The overall performances of the proposed system were evaluated by comparing the recovered trajectories with the real on-line counterpart trajectories. The experiments aim to answer the following questions:
• Q1: Are the clusters correctly solved?
• Q2: Are the components correctly detected in the images?
• Q3: Does the proposed system recover the trajectories in a correct order? We used the complete SigComp2009, SUSIG-Visual and SVC-Task2 as third-party databases for the experiments (see appendix).

### A. Used Metrics

The evaluation is carried out at the pixel level between real and recovered 8-connected trajectories. To this aim, the on-line data of a specimen were interpolated to generate an 8-connected trajectory through Bresenham's line drawing algorithm [65] without any further processing [45]. As metrics, we used the Root Mean Square Error (RMSE) [48], Signal-to-Noise-Ratio (SNR) [66] and Dynamic Time Warping (DTW) [67] to quantify the matching between them. These metrics are defined as follows:

$$\text{RMSE} = \sqrt{\frac{1}{n}\left(\sum_{i=1}^{n}(x_i - \hat{x}_i)^2 + \sum_{i=1}^{n}(y_i - \hat{y}_i)^2\right)} \tag{4}$$

$$\text{SNR} = 10\log\left(\frac{\sum_{i=1}^{n}\left((x_i - \bar{x}_i)^2 + (y_i - \bar{y}_i)^2\right)}{\sum_{i=1}^{n}\left((x_i - \hat{x}_i)^2 + (y_i - \hat{y}_i)^2\right)}\right) \tag{5}$$

$$d(n,n) = \text{DTW}\left(\sum_{i=1}^{n}\sum_{j=1}^{n}\sqrt{(x_i - \hat{x}_j)^2 + (y_i - \hat{y}_j)^2}\right) \tag{6}$$

where $(x, y)$ and $(\hat{x}, \hat{y})$ are the points belonging to the real and recovered trajectories, respectively. For the sake of comparison [14], the real and estimated recovered trajectories are normalized by using a cubic spline, where its length, $n$, was the total sampling points in the real on-line trajectory. Next, min-max scaling is worked out.
It follows from the definition that the smaller the RMSE and DTW, the more similar the real and recovered trajectories. Conversely, higher values of the SNR represent a higher similarity between trajectories.

Furthermore, we investigate the relation between the performance and the complexity (ℂ) of the specimens, since the more complex the handwriting, the more difficult the reconstruction of its writing order. Accordingly, we define the complexity as:

$$\mathbb{C} = \alpha_1 \cdot n_c + \alpha_2 \cdot n_{r=3} + \alpha_3 \cdot n_{r>3} \tag{7}$$

where $n_c$ denotes the number of components in the real hand- writing, $n_{r=3}$ the number of 3-rank clusters and $n_{r>3}$ the number of clusters with a rank greater than three. These three factors represent the difficulty in order recovering trajectories, with the number of components being the most critical factor. Accordingly, their coefficients are empirically adjusted as: $[\alpha_1, \alpha_2, \alpha_3] = [0.6, 0.3, 0.1]$.

### B. Accuracy of Cluster Resolution (Q1)

Fig. 9b shows that our system detected 183, 877 out of 186, 165 clusters of 3 and 4-rank in all databases. These two cluster types thus represent 98.77 % of the cases handled. The accuracy rate obtained was over 97.97 % on average for the three datasets. As seen in Fig. 9a, for 3-rank clusters, the lowest and highest rates were 96.15 % for SVCTask2 and 99.26 % for SUSIG-Visual, respectively. As mentioned in Section III, 3-rank clusters are the most challenging to solve, as they may exhibit configurations that are very similar to those of the branches attached to them. Achieving an accuracy rate greater than 96.00 % on them thus is a remarkable feat. These results show that the criteria we designed for pairing the cluster branches capture essential pieces of knowledge about human trajectory execution. They also show that the higher the cluster rank, the lower the accuracy rate. This is expected since recovering correcting higher-rank clusters is more difficult than correcting those that are lower-ranked. We also observe that there is room for improvements for branch pairings of clusters with ranks greater than 6. However, this drawback has only a limited impact on the performance because such clusters are present in less than 2.5 % of the total number of samples.

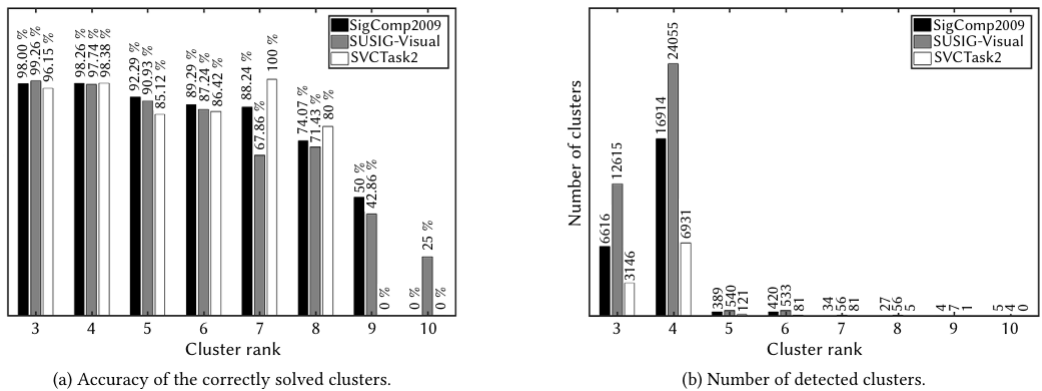

(a) Accuracy of the correctly solved clusters.          (b) Number of detected clusters.

*Fig. 9. Performance of the system solving the clusters found in each database.*

Overall, we have a global accuracy of $\theta = 98.72$ % on SigComp2009, $\theta = 98.91$ % on SUSIG-Visual, and $\theta = 98.59$ % on SVCTask2, which are little better than the results obtained on the SigComp2009 and SUSIG-Visual datasets [50].

### C. Estimation of the Number of Components (Q2)

We assess the estimation of the number of components. Each component has two end-points, corresponding to the points where the pen-tip touches/leaves the tablet. They are used to quantify the number of components in a sample. Therefore, we compare the number of components found by our method with the actual number of components of the on-line samples. Fig. 10 shows the density functions of the real and estimated numbers of components in the samples. We also quantify the density function similarities in terms of the Area Between Curves (ABC). The more similar the density functions, the smaller the ABC value, and therefore, the better the estimation. For all datasets, we obtain excellent performance, but in the case of the SVCTask2, it is clearly outstanding. It is explained since the other two databases contain more complicated and lengthy specimens than the SVCTask2.

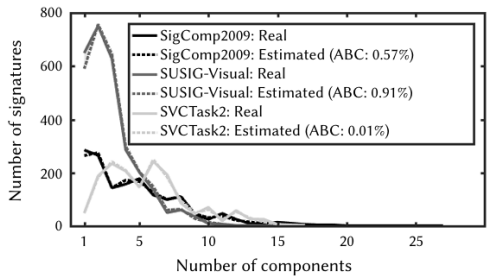

Fig. 10. Density functions of the number of components calculated with real trajectories from different databases and the estimated number of components with our system. ABC denotes the Area Between Curves.

### D. Matching Between Real and Recovered Trajectories (Q3)

This experiment assesses the performance of our system when the order of a static trajectory is wholly recovered. One of the main factors influencing the complete order recovering is the selection of the starting point of each component. To evaluate the extent to which the selection of the starting points determines the recovery trajectories, we gradually relax the condition of the static samples by adding some information regarding the ending points: 1) the Estimated Starting point Nearest Criterion (ESTNC), which selects as the starting point of the next component the nearest untraced estimated starting point; 2) the Real Starting/Ending point Nearest Criterion (RSENC), which provides the system with the coordinates of the real starting and ending points of the components, and 3) the Real Starting/Ending point Ordered Criterion (RSEOC), which provides the system with the coordinates of the real starting and ending points, as well as the correct order of these coordinates. By comparing the results of the third and second scenarios, it is possible to estimate the efficacy of the criterion for finding the starting point of the next component. As the first scenario represents the operating condition of our method, comparing its performance with the second one allows evaluating the efficacy of the criterion for selecting the starting point of each component.

TABLE V. Overall Performance Results When Complete Trajectories Are Recovered in Static Trajectories in Terms of RMSE, SNR, and DTW

| Dataset: SigComp2009 | | Low complexity | Medium complexity | High complexity | Total |
|---|---|---|---|---|---|
| SNR | ESTNC | 28.93 ± 1.26 | 8.15 ± 0.63 | 5.93 ± 0.50 | 14.39 ± 0.56 |
| | RSENC | 33.05 ± 1.25 | 10.70 ± 0.73 | 6.06 ± 0.55 | 16.65 ± 0.60 |
| | RSEOC | 41.16 ± 1.07 | 26.63 ± 0.72 | 14.63 ± 0.64 | 27.50 ± 0.55 |
| RMSE | ESTNC | 0.14 ± 0.01 | 0.27 ± 0.01 | 0.27 ± 0.01 | 0.22 ± 0.00 |
| | RSENC | 0.11 ± 0.01 | 0.25 ± 0.01 | 0.29 ± 0.01 | 0.22 ± 0.01 |
| | RSEOC | 0.04 ± 0.00 | 0.07 ± 0.01 | 0.16 ± 0.01 | 0.09 ± 0.00 |
| DTW | ESTNC | 2.73 ± 0.18 | 6.51 ± 0.22 | 8.82 ± 0.30 | 6.01 ± 0.15 |
| | RSENC | 2.05 ± 0.15 | 6.16 ± 0.23 | 9.39 ± 0.31 | 5.86 ± 0.16 |
| | RSEOC | 0.71 ± 0.09 | 1.82 ± 0.14 | 5.42 ± 0.29 | 2.65 ± 0.12 |

| Dataset: SUSIG - Visual | | Low complexity | Medium complexity | High complexity | Total |
|---|---|---|---|---|---|
| SNR | ESTNC | 23.74 ± 0.83 | 14.35 ± 0.64 | 7.75 ± 0.51 | 15.40 ± 0.41 |
| | RSENC | 32.32 ± 0.79 | 19.94 ± 0.69 | 9.45 ± 0.57 | 20.73 ± 0.44 |
| | RSEOC | 43.34 ± 0.63 | 32.70 ± 0.59 | 19.50 ± 0.62 | 31.98 ± 0.40 |
| RMSE | ESTNC | 0.19 ± 0.01 | 0.23 ± 0.01 | 0.28 ± 0.01 | 0.23 ± 0.00 |
| | RSENC | 0.11 ± 0.01 | 0.18 ± 0.01 | 0.27 ± 0.01 | 0.19 ± 0.00 |
| | RSEOC | 0.03 ± 0.00 | 0.07 ± 0.00 | 0.16 ± 0.01 | 0.08 ± 0.00 |
| DTW | ESTNC | 2.71 ± 0.10 | 3.70 ± 0.12 | 5.33 ± 0.13 | 3.90 ± 0.07 |
| | RSENC | 1.58 ± 0.09 | 3.00 ± 0.12 | 5.33 ± 0.14 | 3.28 ± 0.07 |
| | RSEOC | 0.36 ± 0.04 | 1.13 ± 0.08 | 3.05 ± 0.12 | 1.51 ± 0.05 |

| Dataset: SVCTask2 | | Low complexity | Medium complexity | High complexity | Total |
|---|---|---|---|---|---|
| SNR | ESTNC | 23.72 ± 0.84 | 7.14 ± 0.54 | 2.62 ± 0.25 | 11.19 ± 0.41 |
| | RSENC | 31.61 ± 0.83 | 8.70 ± 0.65 | 2.11 ± 0.34 | 14.18 ± 0.49 |
| | RSEOC | 37.39 ± 0.63 | 29.30 ± 0.69 | 20.43 ± 0.60 | 29.08 ± 0.41 |
| RMSE | ESTNC | 0.13 ± 0.01 | 0.27 ± 0.01 | 0.31 ± 0.01 | 0.24 ± 0.00 |
| | RSENC | 0.09 ± 0.01 | 0.28 ± 0.01 | 0.35 ± 0.01 | 0.24 ± 0.01 |
| | RSEOC | 0.03 ± 0.00 | 0.06 ± 0.00 | 0.09 ± 0.00 | 0.06 ± 0.00 |
| DTW | ESTNC | 1.72 ± 0.11 | 3.62 ± 0.11 | 4.97 ± 0.10 | 3.43 ± 0.07 |
| | RSENC | 1.16 ± 0.10 | 3.74 ± 0.12 | 5.69 ± 0.11 | 3.52 ± 0.08 |
| | RSEOC | 0.34 ± 0.04 | 0.79 ± 0.07 | 1.46 ± 0.08 | 0.86 ± 0.04 |

In Table V, for each database, scenario and level of complexity, we list the performance in terms of the mean and standard error of the performance measures mentioned above.

As expected, we can observe that the simpler the handwriting, the better the reconstruction, for all metrics. They also show that in case of a low complexity handwriting, the performance is very similar across the three datasets, but diverges more and more as the handwriting complexity increases.

With regards to the metrics, the performances in Table V show that the SNR is more sensitive than RMSE and DTW to the errors in recovered trajectories. Nevertheless, all metrics maintain a similar range of values in all cases.

Last but not least, the results in the table show that a significant improvement in the performance is obtained when both the starting/ ending points of each component and their orders are made available to our method. This observation is independent of the database and the level of complexity[3].

*TABLE VI. Performance Comparison of Recovered Trajectories of the Estimated STarting Point Nearest Criterion (ESTNC) With Other Works*

| Paper | Dataset | RMSE* | DTW* |
|---|---|---|---|
| – *Related works about recovering words* – | | | |
| [48] | Private - Single Strokes (on-line transformed in off-line) | 400.98 | 118.12 |
| [48] | Private - Multi Strokes (on-line transformed in offline) | 503.81 | 171.91 |
| [48] | Private - Scanned Words | 2538.05 | 553.57 |
| [48] | IRONOFF [49] | 669.03 | 278.27 |
| [56] (*LSTM*) | Unipen | - | 0.04 |
| [56] (*Conv*) | Unipen | - | 0.04 |
| [56] (*Class Avg*) | Unipen | - | 0.21 |
| – *Related works about recovering signatures* – | | | |
| [14] (*Best system on DTW*) | Public - Arabic signature [33] | 0.34 | 28.81 |
| [14] (*Best system RMSE*) | Public - Arabic signature [33] | 0.25 | 52.40 |
| [50] | SigComp2009 | 0.06 | 382.01 |
| [50] | SUSIG-Visual | 0.05 | 300.50 |
| **This work** | **SigComp2009** | **0.22** | **6.01** |
| | **SUSIG-Visual** | **0.23** | **3.90** |
| | **SVCTask2** | **0.24** | **3.43** |

\* Note that different formulas for RMSE and DTW are used in the papers, making a fair comparison more difficult. We used the formulas proposed in the competition presented in [14].

---

[3] A video showing the estimated recovered trajectories with different levels of complexity is available at https://youtu.be/TYoZZ8CThhw.

To put our results in context, Table VI shows the performances obtained in related works. We can see different performance ranges among the works, suggesting that no standard procedure has thus far been established for measuring the effectiveness of the writing order recovery. For this reason, beyond the implementation of RMSE and DTW metrics, it is necessary to take into account data normalization, data aggregation, and, above all, the handwriting database used, whose complexity is not easily measurable. Nonetheless, the results of such comparisons are useful, as they generally convey a rough estimate of advancements in the field, even though it does not provide a fair basis of comparison.

## 6. Conclusions and Outlook

We have developed a system for recovering the ballistic trajectory order of long and complex thinned static handwritten signatures. Our system operates in three stages: (*i*) point classification, (*ii*) local examination, and (*iii*) global reconstruction. In the point classification, the clusters are identified and correspond to the agglomeration of pixels in the images. The agglomerations of lines correspond to crossings of lines in the thinned trajectories. Thus, a cluster can be characterized by the number of input-output lines or branches. In the local examination, input-output branches are paired by exploiting heuristic rules inspired by both good continuity and motor control principles when signing, with preference given to smooth and straight ballistic trajectories. At the global reconstruction stage, the end-points of the components (pendowns) are identified and sorted. Once a component is re-covered, the system decides on the new component to recovering.

This procedure requires that a number of parameters and weights be adjusted. Their values are determined heuristically by trial and error, as the best matching between the real and reconstructed trajectory is sought. Furthermore, both the sensitivity and stability of the results with respect to these parameters are studied, and we see that the performance of our procedure is barely affected by a variation of up to 10 % of these parameters. To avoid overfitting in these values and make the results more meaningful, the parameters are adjusted with a subset of the SigComp2009 signature database. Then, the results are obtained with different publicly available databases, namely, the complete SigComp2009, SUSIG-Visual and SVCTask2.

The performance of the system is analyzed considering several aspects. As the ground truth of our experiments, we use the on-line trajectories, which contain details of how real signers wrote the trajectory. We first observe a competitive performance when the branches are paired on the clusters. It is worth pointing out that the branch association in the clusters is the first step towards the final writing order recovery. Moreover, a few mistakes may lead to an overall error in the estimation of trajectory order. Secondly, we also study the number of components estimated in the signature and the complete trajectory order recovery. For research purposes, our system can be freely downloaded from GitHub.

Although promising results are observed in this work, more efforts are required to ensure a more reliable estimation of the trajectories. Our experiments suggest that it is of paramount importance to continue investigating the rules for choosing the ending points of the components and their order, taking into account unknown pen-ups trajectories. Indeed, having the availability of on-line trajectories to recover a static one could improve both the cluster resolution and the complete order recovering. In this case, mapping two skeleton-based images is a further strategy that could be explored by using optical flow analysis, diffeomorphism functions, or inkball models, among others.

In a challenging framework, which uses an off-line handwriting as input and approximates its corresponding on-line counter- part as output, i.e. $(x(t), y(t))$, our system plays an important role. Nevertheless, this framework also implies that more effort is required in thinning algorithms to improve the handwriting image quality and resolution. Furthermore, estimating temporal properties in the recovered trajectories is another open question. A possible solution is to assign a timestamp sequence to the 8-connected trajectories. This would open the door to working out dynamic properties such as the velocity or acceleration. In the meantime, the proposed system constitutes a reasonable starting point for future research in the challenging field of on-line trajectory estimation from off-line specimens.

## Appendix

### A. Databases

We evaluate the proposed system on signatures because they represent long and complex handwriting, contain text and flourishes, and their patterns result in multiple pen-downs and a high number of clusters, so constituting a suitable and very challenging benchmark.

We use the following data of three publicly databases[4]:
- SigComp2009 [64]. Contains 1552 on-line Western signatures written by 79 subjects. There are 932 genuine signatures available since each signer gave 12 specimens on average.

- SUSIG-Visual corpus [68]. Includes 2820 on-line Western signatures written by 94 subjects. Each participant produced 20 genuine signatures in two sessions, i.e., $94 \times 20 = 1880$ specimens.
- SVCTask2 [69]. Consists of 1600 signatures written by 40 subjects, 17 of whom used Oriental scripts and 23, Western scripts. Each subject produced 20 genuine signatures.

SigComp2009 and SUSIG-Visual, which are made up of only Western signatures, were used in a bid to evaluate the independence of the proposed method from the database, whereas the use of SVCTask2, also composed of Oriental specimens, was intended to show the independence of the algorithm from the script type.

Eventually, the skeleton of the off-line handwriting was obtained by converting the on-line trajectories into 8-connected, one-pixel- wide digital lines through the Bresenham's line drawing algorithm [65] without any further processing [45]. The out-put resolution of the images was 600 dpi. This choice, moreover, provides a perfect spatial matching between the on-line and the off-line representations of the trajectories, establishing a solid ground truth for performance evaluation.

---